\pdfoutput=1

\documentclass[11pt]{article}

\usepackage[]{acl}

\usepackage{times}
\usepackage{latexsym}

\usepackage[T1]{fontenc}

\usepackage[utf8]{inputenc}

\usepackage{microtype}

\usepackage{graphicx}
\usepackage{amssymb} 

\usepackage{footmisc}

\title{Resolving References in Visually-Grounded Dialogue via Text Generation}

\author{
    Bram Willemsen \and Livia Qian \and Gabriel Skantze \\
    Division of Speech, Music and Hearing \\ KTH Royal Institute of Technology \\ Stockholm, Sweden \\
    \{\texttt{bramw}, \texttt{liviaq}, \texttt{skantze}\}\texttt{@kth.se}
}

\begin{document}
\maketitle
\begin{abstract}
Vision-language models (VLMs) have shown to be effective at image retrieval based on simple text queries, but text-image retrieval based on conversational input remains a challenge. Consequently, if we want to use VLMs for reference resolution in visually-grounded dialogue, the discourse processing capabilities of these models need to be augmented. To address this issue, we propose fine-tuning a causal large language model (LLM) to generate definite descriptions that summarize coreferential information found in the linguistic context of references. We then use a pretrained VLM to identify referents based on the generated descriptions, zero-shot. We evaluate our approach on a manually annotated dataset of visually-grounded dialogues and achieve results that, on average, exceed the performance of the baselines we compare against. Furthermore, we find that using referent descriptions based on larger context windows has the potential to yield higher returns.

\end{abstract}

\section{Introduction}
\begin{figure}
    \centering
    \includegraphics[width=0.9\columnwidth]{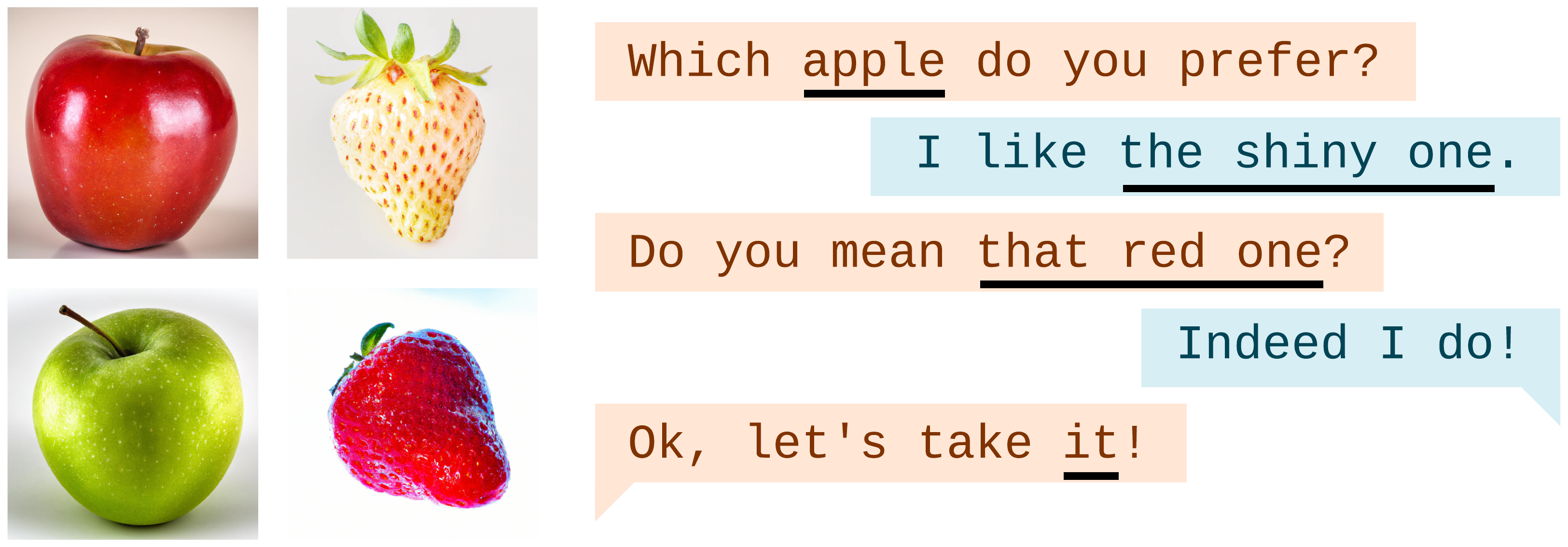}
    \caption{Example dialogue in which two participants discuss fruits. Expressions that denote one or more images are underlined.}
    \label{fig:example}
\end{figure}

Visually-grounded dialogues are conversations in which participants make references to the visual world.
Referring in conversation is understood to be a collaborative process, with shared responsibility for ensuring the successful identification of the referent \citep{clark_referring_1986}. 
It is not uncommon for a definite reference to be established over multiple turns, with each separate contribution unlikely to be a minimally distinguishable description of the referent.
Taken out of their use context, these referring expressions may be difficult, if not impossible, to resolve. 
Consider the example dialogue in Figure \ref{fig:example}. 
The underspecified description \textit{``the shiny one''} leads to a clarification question, \textit{``Do you mean that red one?''}.
To resolve the expression \textit{``that red one''} to its referent, we need information from earlier in the conversation to understand that \textit{``one''} is a proform of \textit{``apple''}.
Without this linguistic context, the red strawberry and the red apple are equally likely referents.

We can break the problem of reference resolution in visually-grounded dialogue down into three subproblems: (1) mention detection, or finding the expressions that can be grounded in the visual context (\textit{``that red one''}); (2) aggregation of referent-specific information (linking \textit{``apple''}, \textit{``the shiny one''}, and \textit{``that red one''}); and (3) referent identification, or the grounding of language (finding the referent that is best described by the three expressions from among a set of candidate referents). 
This final step requires bridging the gap between vision and language. 
For this purpose, we can turn to pretrained vision-language models (VLMs), which have shown to be effective at zero-shot text-image retrieval when given a description of an image \citep[e.g.,][]{radford_learning_2021,jia_scaling_2021,li_blip-2_2023}. 
However, current VLMs lack the discourse processing capabilities necessary for reference resolution in visually-grounded dialogue. 
Although some VLMs may correctly identify the red apple as the referent given the entire dialogue of Figure \ref{fig:example}, dialogues are often vastly more complex than this hypothetical exchange. Take, for instance, the dialogue in Appendix \ref{sec:appendix-dialogue-excerpt}: with multiple mentions of different referents within the same utterance, such a brute-force method would immediately fail. It is clear that if we want VLMs to be effective for this purpose, their discourse processing capabilities need to be augmented.

To this end, we propose fine-tuning a causal large language model (LLM) for the task of \textit{referent description generation}. 
Referent description generation can be regarded as a special case of referring expression generation with the goal of always generating the most complete expression possible. 
For a given mention, the model is trained to generate a definite description that summarizes all information that has been explicitly disclosed about the referent during a conversation. For example, for the mention \textit{``that red one''} in Figure \ref{fig:example} we would want the model to generate the description \textit{``the shiny red apple''}.
We will refer to the fine-tuned model as the \textit{conversational referent description generator} (CRDG). The description generated by the CRDG is then used by a pretrained VLM to identify the referent, zero-shot. 
Our approach can be seen as an exploration of the limits of depending on linguistic context alone for generating referent descriptions, as the discourse processing and eventual grounding of the descriptions are entirely disjoint.

For the experiments presented in this paper we use data from the collaborative image ranking task A Game Of Sorts \citep{willemsen_collecting_2022}. 
Referents are represented by separate, but visually similar images from a shared entity category. Due to their largely unrestricted nature and with a focus on the collaborative referential process, the collected dialogues form a challenging test bed for visually-grounded language understanding in conversation. 
We manually annotate the dialogues by marking mention spans and aligning the spans with the images they denote, and provide both manually constructed and automatically derived ``ground truth'' referent descriptions based on our manual annotations for all marked mentions.

Our main contributions are as follows:
\begin{itemize}
    \item We present a generative approach to reference resolution in visually-grounded dialogue that frames the discourse processing side of the task as a causal language modeling problem;
    \item We show that it is possible to fine-tune a causal LLM to generate referent descriptions from dialogue to be used by a pretrained VLM for referent identification, zero-shot;
    \item We release the discussed materials, including our annotations for A Game Of Sorts \citep{willemsen_collecting_2022}\footnote{\url{https://github.com/willemsenbram/reference-resolution-via-text-generation}, \href{https://doi.org/10.5281/zenodo.8176114}{doi:10.5281/zenodo.8176114} \label{fn:zenodo}}.
\end{itemize}

\section{Background}
Visually-grounded language understanding is fundamental for conversational agents that engage in dialogue involving references to the visual world.
Researchers have introduced a variety of tasks that provide data for development and frameworks for evaluation of visually-grounded dialogue models.
These tasks often take the form of goal-oriented, dyadic interactions but differ in terms of, for example, the visual stimuli used, e.g. abstract figures or realistic photos; the roles assigned to participants, e.g. whether symmetric or asymmetric; constraints on message content, e.g. a fixed vocabulary; and the nature of the task, e.g. navigation, identification, ranking, and multi-turn visual question answering \citep[e.g.][]{das_visual_2017,de_vries_guesswhat_2017,shore_kth_2018,ilinykh_meet_2019,haber_photobook_2019,udagawa_natural_2019,willemsen_collecting_2022}.
It has been noted that the task configuration can significantly impact the extent to which certain dialogue phenomena, such as coreferences and clarification requests, are represented in the collected data, if at all \citep{agarwal_history_2020,haber_photobook_2019,ilinykh_meet_2019,schlangen_grounded_2019,willemsen_collecting_2022}. 
Tasks that heavily constrain the interactions do not reflect the complex nature of dialogue to the same degree as tasks that have been designed for these phenomena to naturally emerge as part of the discourse, such as A Game Of Sorts \citep{willemsen_collecting_2022}, which we use in this paper.

The terms referring expression comprehension \citep[e.g.][]{yu_modeling_2016}, referring expression grounding \citep[e.g.][]{zhang_grounding_2018}, referring expression recognition \citep[e.g.][]{cirik_visual_2018}, and reference resolution \citep[e.g.][]{kennington_discriminative_2015} have been used interchangeably to describe the problem of mapping the language that denotes a referent to a representation of that referent in the visual modality.
Prior work noted the importance of referring expressions to conversation, but often modeled the problem independent of the dialogue \citep[e.g.][]{cirik_visual_2018,schlangen_resolving_2016,yu_modeling_2016,zhang_grounding_2018}. 
The granularity at which grounding occurs may differ between works, as the language may be mapped to bounding boxes of individual objects \citep{cirik_visual_2018,schlangen_resolving_2016,yu_modeling_2016,zhang_grounding_2018}, objects or larger image regions represented by segmentation masks \citep{liu_recurrent_2017}, or entire images altogether \citep{haber_photobook_2019,takmaz_refer_2020}.

To address the problem computationally, both modalities must in some way be encoded. 
Engineered visual feature representations and simple language models such as those based on n-grams \citep[e.g.][]{kennington_discriminative_2015,kennington_simple_2017,shore_using_2018} have been mostly replaced with more powerful learned representations that embed the images and text in high-dimensional vector spaces \citep{haber_photobook_2019,takmaz_refer_2020}. 
This has made it possible to resolve references by computing representational similarity between an encoding of the text that contains a mention and the embeddings of the candidate referents, where the candidate that has the highest matching score is assumed to be the referent \citep{haber_photobook_2019,takmaz_refer_2020}.

Recent work on multimodal representation learning has shown that jointly embedding text and images can work at scale. Trained using a contrastive objective, maximizing representational similarity between true pairings of images and text while simultaneously minimizing similarity of false pairs, vision-language models (VLMs) such as CLIP \citep{radford_learning_2021}, ALIGN \citep{jia_scaling_2021}, BLIP \citep{li_blip_2022}, and BLIP-2 \citep{li_blip-2_2023}, have shown to be effective zero-shot classifiers, outperforming the previous state-of-the-art on various benchmarks without the need for further fine-tuning on specific tasks. 
However, despite their noteworthy image-text matching performance based on simple text queries, these VLMs lack the discourse processing capabilities required for reference resolution in visually-grounded dialogue. 
Even a simplified example, such as shown in Figure \ref{fig:example}, illustrates a fundamental challenge, namely that of coreference resolution.
The interpretation of anaphoric pronouns, such as \textit{``it''}, is dependent on their antecedents. Without resolving its coreferences first, identifying the referent based on the pronoun alone leads to a random guess. 

To improve downstream performance on discourse processing tasks involving coreference, prior work has approached the problem as one of transforming the original input based on linguistic context. This was done either via substitution, such as in \citet{bhattacharjee_investigating_2020} where pronouns were substituted for more descriptive mentions of the same referent, or via generation, such as in \citet{quan_gecor_2019} where entire utterances were reconstructed in a pragmatically complete manner with coreferences and ellipses resolved. To the best of our knowledge, this approach has not yet been applied to reference resolution in visually-grounded dialogue.

Most contemporary natural language processing (NLP) works use Transformer-based language models \citep{vaswani_attention_2017}. For text generation tasks, it is common to use (unidirectional) autoregressive, or \textit{causal}, language models such as GPT \citep{radford_improving_2018}. While processing sequences, causal language models mask the future, allowing the model to only attend to the current and previous tokens while predicting the next token. 
A persistent trend has been to scale up language models, both in terms of their parameter count and the size of their training datasets.
These increasingly larger models, such as GPT-3 \citep{brown_language_2020}, OPT \citep{zhang_opt_2022}, PaLM \citep{chowdhery_palm_2022}, 
and LLaMa \citep{touvron_llama_2023}, have been dubbed \textit{large language models} (LLMs).
The current leading paradigm to modeling downstream NLP tasks is based on transfer learning, where a pretrained LLM is fine-tuned for a specific task on a smaller, domain-specific dataset.

\section{Method}

\begin{figure*}[t!]
    \centering
    \includegraphics[width=1\textwidth, trim={0 0 0 1cm},clip]{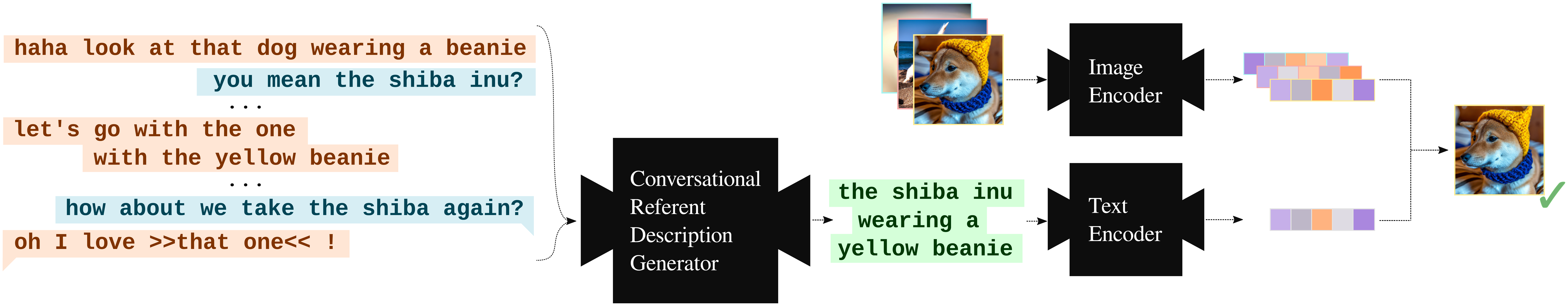}
    \caption{The proposed visually-grounded reference resolution framework. With the CRDG we generate a referent description for a marked mention, to be used by a (frozen) pretrained VLM for referent identification.
    }
    \label{fig:model}
\end{figure*}

We treat visually-grounded reference resolution as a text-image retrieval task, where referents are represented by images.
We leave finer-grained grounding of words and phrases to image regions or individual entities or parts thereof for future work.

\subsection{Proposed Framework}
We frame the discourse processing side of the task as a causal language modeling problem. 
Figure \ref{fig:model} shows a visualization of the proposed framework.

\noindent \textbf{Task Definition} We denote the dialogue as $D = (u_1, u_2, ..., u_n)$, where each $u_i$ represents an utterance. 
Each utterance consists of an ordered sequence of tokens.
An utterance may contain one or more mentions, denoted as $M$.
A mention is an ordered subsequence of tokens from an utterance.
A mention has an exophoric referent, denoted as $R$.
A mention is embedded in what we call its linguistic context, denoted as $L$. 
As an ordered subsequence of $D$, the linguistic context of a given mention consists of the utterance in which it is contained and all preceding utterances. 
The number of preceding utterances, hereafter referred to as the dialogue history, may be capped if a finite size context window is defined. 
The aim of visually-grounded reference resolution is to resolve a reference to its referent, i.e. to identify $R$ for a given $M$, from a set of candidate referents, denoted as $C$, such that $R \subseteq C$; $|R| = 1$ for single-image referents, $|R| > 1$ for multi-image referents, and $R = C$ if $M$ refers to all members in $C$.

\noindent \textbf{Referent Description Generation} We propose to generate a definite description, denoted as $Y$, for a given mention $M$ that summarizes all that has been disclosed in $L$ about the referent $R$. 
For this purpose, we fine-tune a causal LLM that learns to generate $Y$ conditioned on $L$.
$Y$ is a sequence of tokens expected to be largely constructed from tokens that appear, or are some derivative of tokens that appear, in the coreference chain of $R$, which is contained in $L$.
We refer to the fine-tuned model as the \textit{conversational referent description generator} (CRDG).
For an example of the context dependency of referent description content, see Figure \ref{fig:label_example} in Appendix \ref{sec:appendix-model-input}.

\noindent \textbf{LLM Input} We mark $M$ in $u_i$ by inserting positional markers as special tokens to indicate the beginning and end of the mention span.
We prepend each utterance in $L$ with a speaker token to indicate the source of the contribution.
When $D$ is task-oriented, we update $L$ by prepending task instructions, i.e. a special token followed by a sequence of tokens describing the task performed by the dialogue participants. 
For an example of the input to the LLM, see Figure \ref{fig:model_input} in Appendix \ref{sec:appendix-model-input}.

\noindent \textbf{Text-Image Retrieval} We use a pretrained VLM to identify $R$ from $C$ based on $Y$, zero-shot.
We use the text encoder of the VLM to encode $Y$ into an $n$-dimensional feature vector, denoted as $\mathbf{v}$. We use the image encoder of the VLM to encode each candidate referent of $C$ into an $n$-dimensional feature vector, which gives a $|C| \times n$ matrix, denoted as $\mathbf{A}$. 
We then compute their matrix-vector product. 
For single-image referents, i.e. when $|R| = 1$, we take the referent to be $R = argmax(\mathbf{A}\mathbf{v})$.

In order to produce accurate referent descriptions, the CRDG must implicitly learn to perform coreference resolution as we do not provide explicit supervision for this subtask. In each sample, only the current mention for which we want the model to generate a description is marked; none of its coreferences are in any way indicated.
A principal advantage of our model is that it can resolve multiple mentions, even when they have different referents, appearing in the same utterance, including nested mentions.
Note that for the purpose of this study, we assume mention detection to be solved. As it stands, using this framework in production requires a separate model to propose candidate mentions at the span level.

\subsection{Baseline Models}
As a lower bound, we report random chance performance. 
In addition, we compare performance of our approach to baselines based on simple heuristics and a coreference resolution model. 

\subsubsection{Heuristics}
\textbf{Mention}
We evaluate the image retrieval performance when the VLMs are presented with just the marked mentions.

\noindent \textbf{Substitution}
We improve upon the mention-only baseline by substituting proforms, e.g. pronouns such as \textit{``it''}, and mentions without descriptive content, e.g. phrases such as \textit{``the one you mentioned''}, with the most recent mention that does not belong to either category. 
This is expected to be a relatively strong baseline when mentions are specific and anaphora have mostly local antecedents. 

\subsubsection{Coreference Resolution}
We opt for an off-the-shelf\footnote{\url{https://github.com/allenai/allennlp-models/tree/main/allennlp_models/coref}} span-based coreference resolution model (\textbf{coref}) originally presented in \citet{lee_higher-order_2018}, but that has since been updated to use SpanBERT \citep{joshi_spanbert_2020} instead of the original GloVe embeddings \citep{pennington_glove_2014}.
For each mention, we use the model to resolve its coreference links and aggregate all coreferential information in its cluster based on the given context window. 

We experiment with two different representations of the referent descriptions from this model, those being (1) a concatenation of all of the mention's coreferences and (2) an ordered \textit{set-of-words} representation that contains only the unique lexical items in the cluster.
To offset that this model was not specifically trained to handle coreference in conversation, we provide it with the contents of the span of the mention when it does not manage to detect the mention itself and, consequently, not connect it to any of its coreferences.
For partial matches, in addition to adding all tokens from the cluster associated with the match, we also add the missing tokens from the span to the description.

\section{Experiments}

\subsection{Data}
We use the dialogues from the collaborative image ranking task \textbf{A Game Of Sorts} \citep[AGOS,][]{willemsen_collecting_2022} for our experiments.
In AGOS, two players are asked to rank a set of images based on a given sorting criterion. 
They see the same set of images, but the position of the images on the screen is randomized for each player.
Through a largely unrestricted conversation, and without being able to see the perspective of the other player, the players need to agree on how to rank the images given the sorting criterion.
Sorting criteria are embedded in scenarios that are intended to create a discussion, leading to mixed-initiative interactions with both parties contributing to the discourse. Each interaction takes place over four rounds with the same set of nine images, effectively guaranteeing repeated references.
The image sets used for the game cover five different image categories. 
Each set contains nine images with each image representing an entity from one of these categories as its main subject. \citet{willemsen_collecting_2022} collected three interactions per image set for a total of 15 dialogues.

\noindent\textbf{Ground Truth}
Our formulation of the visually-grounded reference resolution problem requires span-based annotations of mentions aligned with the image(s) they denote. These annotations are the basis of what we will refer to as our ``ground truth'' references used for both training and evaluation.
We follow \citet{willemsen_collecting_2022} regarding the marking of mentions in AGOS, in that we only annotate those that are either singletons or are part of an identity relation with other mentions that have an exophoric referent that is part of the visual context, i.e. regardless of form, any referring expression that is meant to denote one or more of the images.
During the game, players were asked to provide self-annotations: for each message they sent they were asked to indicate which image(s), if any, they were referring to. We use these self-annotations, post-edited where necessary, to manually mark the spans of mentions that can be grounded in the visual context.

We create three different representations of the ``ground truth'' referent descriptions.
Two are automatically extracted from the marked mentions and are similar in structure to the labels of the \textbf{coref} baseline, i.e. (1) an incremental concatenation of the reference chain and (2) an incremental ordered set of words consisting of the unique lexical items in the cluster. The third are manually constructed labels that summarize reference chains as definite descriptions.
For each representation, the context window dictates which references are considered for the label. 

\subsection{Model Specifications}
For pointers to implementations, we refer the reader to our repository\footref{fn:zenodo}.

\subsubsection{LLMs}
We fine-tune two LLMs, GPT-2 \citep{radford_language_2019} and GPT-3 \citep{brown_language_2020}, for conversational referent description generation. For hyperparameters, see our Supplementary Material. 

\noindent \textbf{GPT-2} We fine-tune the 1.5 billion parameter GPT-2 model. 

\noindent \textbf{GPT-3} We fine-tune the 175 billion parameter \texttt{davinci} base model using the OpenAI API. 

\subsubsection{VLMs}
We evaluate the zero-shot text-image retrieval performance of several pretrained VLMs for our task, those being CLIP \citep{radford_learning_2021}, ALIGN \citep{jia_scaling_2021}, BLIP \citep{li_blip_2022}, and BLIP-2 \citep{li_blip-2_2023}. 

\noindent \textbf{CLIP} We evaluate two variants of CLIP, CLIP ViT-B/32 and CLIP ViT-L/14. 

\noindent \textbf{ALIGN} 
We use the COYO-ALIGN implementation trained from scratch on COYO-700M. 

\noindent \textbf{BLIP} We use the BLIP base model. 

\noindent \textbf{BLIP-2} We use the BLIP-2 model that was fine-tuned on the \citet{karpathy_deep_2015} training set split of MS COCO \citep{lin_microsoft_2014}.

\subsection{Evaluation}
We perform (nested) five-fold cross-validation by partitioning the AGOS dataset along the five image sets. To avoid leakage, for each run we use the three dialogues from one image set as the held out test set and train on the twelve dialogues from the four other image sets. To evaluate how dialogue history affects results, we report performance of the different methods for two context windows, $\mathbf{3}$ and $\mathbf{7}$. In addition, we examine whether increasing the size of the context window further would, in principle, lead to greater returns, by assessing ground-truth performance for windows of $\mathbf{13}$ and the $\mathbf{full}$ dialogue context. 
Finally, we conduct an error analysis of the generated descriptions.

Note that because we do not incorporate game state information with respect to the visual context during training, we make a simplifying assumption with regard to the images and reduce the candidate set, at test time, as the game progresses. A successfully ranked image is no longer considered part of the visual context for that round. Although this does mean that the models will not be able to identify the referent for references to ranked images, as they will not be part of the candidate set, such references are an extremely rare occurrence, as players must discuss the unranked images to progress with the task.
For the sake of completeness, we will also report results for the unchanged candidate set. 

\subsubsection{Metrics}
We measure task success for visually-grounded reference resolution in terms of text-image retrieval performance. In addition, we estimate the quality of the generated referent descriptions by comparing them to the manually constructed ground truth labels using text similarity metrics.

\noindent\textbf{Text-Image Retrieval}
We estimate the image retrieval performance based on accuracy $[0, 1]$, mean reciprocal rank (MRR) $[0, 1]$, and normalized discounted cumulative gain (NDCG) $[0, 1]$.
We limit our evaluation to single-image referents. 
Accuracy is top-1 accuracy. 

For our random lower bound, we can calculate the expected values for accuracy and MRR. For top-1 accuracy we take 1 over the size of the set of candidate images per item, averaging over all items. 
For MRR we take 1 over the size of the set of candidate images, divided by two per item, averaging over all items. 
Calculating an expected value for NDCG of a random model is intractable due to its dependence on relevancy scores.

\noindent\textbf{Text Generation}
We evaluate the output from the CRDGs by comparing the generated descriptions to the manually constructed ground truth labels using metrics to quantify similarity. 
We use the Jaccard index $[0, 1]$ to assess vocabulary overlap.
We use BLEU $[0, 1]$ \citep{papineni_bleu_2002} to assess similarity based on n-gram overlap (unigrams to four-grams).
We use the longest common subsequence variant of ROUGE $[0, 1]$ \citep{lin_rouge_2004}, i.e. ROUGE-L, as a further indication of the preservation of word order.
In addition, we opt for an embedding-based metric as a proxy for semantic equivalence between the predicted and reference sequences. 
For this purpose, we compute cosine similarity $[0, 1]$ between text embeddings. 

\subsubsection{Human}
We conduct two different human subject experiments to assess human performance for this task. We provide additional details about the experimental setup in the Supplementary Material.

\noindent\textbf{Independent}
We conduct an experiment aimed at comparing VLM and human performance on the task where every trial is independent. Participants are given a referent description and are asked to select from a set of candidate images the image they believe is best described by the label. The images and labels are presented to the participants independent of the dialogue.
Note that we evaluate with the reduced candidate set. The referent descriptions are the manually constructed ground truth labels based on the $\mathbf{full}$ dialogue context.
To collect data for all labels, ensuring independence of observations, we recruited 354 participants via crowdsourcing. The crowdworkers were financially compensated for their contributions.

\noindent\textbf{Holistic}
We conduct an experiment in which mentions are shown to participants within the context of the dialogue. For each mention, the participants are presented with the dialogue leading up to and including the message which contains the reference. The start and end of the span of the mention that the participant is asked to resolve are visually indicated. For each marked mention, the participant is asked to select which image or images are referenced. 
As they progress with the task, participants will have access to increasingly more of the dialogue history. For each mention the participants are presented with all images, but with a visual indication of their status, i.e. for each image whether the players had managed to successfully rank it at that point in the interaction. 
We recruited 23 participants via crowdsourcing.
For each of the 15 AGOS dialogues we collected data from two different participants. 
Each participant was allowed to provide data for at most one dialogue per image set.
The crowdworkers were financially compensated for their contributions.

\section{Results}

\begin{table}[t]
\footnotesize
\centering
\begin{tabular}{lcc|cc|cc}
\hline
\multicolumn{1}{c}{} & \multicolumn{2}{c}{$\mathbf{Accuracy}$} & \multicolumn{2}{|c|}{$\mathbf{MRR}$} & \multicolumn{2}{c}{$\mathbf{NDCG}$} \\ 
\cline{2-7}
& $\mathbf{3}$ & $\mathbf{7}$  & $\mathbf{3}$ & $\mathbf{7}$  & $\mathbf{3}$ & $\mathbf{7}$  \\ 
\hline
Random & .22 & .22  & .43 & .43 & - & - \\ 
Mention & .59 & .59 & .73 & .73 & .79 & .79 \\ 
Substitution & .68 & .68 & .80 & .80 & .85 & .85 \\ 
\hline
coref, chain & .65 & .66 & .78 & .79 & .83 & .84 \\ 
coref, set & .66 & .66 & .78 & .79 & .84 & .84 \\ 
\hline
GT, chain & .73 & .74 & .83 & .85 & .87 & .88 \\ 
GT, set & .73 & .75 & .84 & .85 & .87 & .89 \\ 
GT, manual & .72 & .74 & .83 & .84 & .87 & .88 \\ 
\hline
GPT-2 & .64 & .60 & .77 & .74 & .83 & .80 \\ 
GPT-3 & .69 & .71 & .81 & .82 & .86 & .86 \\ 
\hline
\end{tabular}
\caption{\label{main-ir-results}
Cross-validated image retrieval performance averaged over five folds for single-image referents. \textit{Note}. 
Scores shown are of VLM that averaged best performance (BLIP-2).
Scores are rounded to the nearest hundredth. GT = ground truth. 
}
\end{table}

\subsection{Text-Image Retrieval} 
Table \ref{main-ir-results} shows, for context windows $\mathbf{3}$ and $\mathbf{7}$, the zero-shot text-image retrieval performance results for the VLM that averaged best performance over the five folds, which was BLIP-2. 
For the text-image retrieval accuracy achieved by the other VLMs, performance on the not reduced candidate set, and accuracy per fold for BLIP-2, see Appendix \ref{sec:appendix-vlm-results}.

As can be seen from the results presented in Table \ref{main-ir-results}, we achieve best performance with a fine-tuned GPT-3 as the CRDG and BLIP-2 for zero-shot text-image retrieval.
In addition to outperforming the baselines, we find that GPT-3 is a more performant discourse processor for this task than GPT-2. This result is consistent between the VLMs.

Results generally show a slight increase in performance when increasing the context window from $\mathbf{3}$ to $\mathbf{7}$. Performance on the ground truth reference descriptions for context windows $\mathbf{13}$ and the $\mathbf{full}$ dialogue shows this trend persists, with BLIP-2 achieving approximately $75\%$ and $83\%$ accuracy, respectively. 
A plot of the performance for the four context windows is shown in Figure \ref{fig:ground_truth_accuracy} in Appendix \ref{sec:appendix-vlm-results}.
This result suggests that the size of the context window may have a significant impact on performance, with an $11\%$ increase in accuracy from $\mathbf{3}$ to $\mathbf{full}$.
Furthermore, the VLMs do not seem overly sensitive to the composition of the referent descriptions, as performance is largely comparable between the automatically generated and the manually constructed ground truth labels.

We find that BLIP-2 is on par with human text-image retrieval performance in terms of top-1 accuracy for the manually constructed ground truth referent descriptions based on the full dialogue history for single-image referents, as our human participants averaged roughly $80\%$ accuracy in the independent setup.
However, when we compare these results with the single-image referent text-image retrieval performance in the holistic setup, we see that the upper bound for this task when references are resolved within the combined linguistic and extralinguistic dialogue context is likely considerably higher as our human participants averaged approximately $91\%$ accuracy (average of best performance per dialogue is roughly $93\%$).

\subsection{Text Generation}
Table \ref{main-tg-results} shows the text generation metric results averaged over the five folds, providing an indication of the extent to which the fine-tuned LLMs managed to generate referent descriptions that approximate the manually constructed ground truth labels. 
We observe that an increase in context window size results in a decrease in scores, which is consistent across metrics. 
Interestingly, we did not find such a decrease with respect to text-image retrieval performance.
We do again find GPT-3 to be more performant than GPT-2, here in terms of approximating the ground truth.

\begin{table}
\footnotesize
\centering
\begin{tabular}{lcc|cc}
\hline
\multicolumn{1}{c}{} & \multicolumn{2}{c}{$\mathbf{GPT}$-$\mathbf{2}$} & \multicolumn{2}{|c}{$\mathbf{GPT}$-$\mathbf{3}$} \\ 
\cline{2-5}
& $\mathbf{3}$ & $\mathbf{7}$ & $\mathbf{3}$ & $\mathbf{7}$ \\ 
\hline
BLEU & .55 & .47 & .75 & .70 \\ 
ROUGE-L & .71 & .65 & .86 & .83 \\ 
Jaccard & .44 & .35 & .70 & .63 \\ 
Cosine & .88 & .85 & .96 & .95 \\ 
\hline
\end{tabular}
\caption{\label{main-tg-results}
Text generation metrics evaluation results averaged over five folds for single-image referents. \textit{Note}. 
Scores are rounded to the nearest hundredth.
}
\end{table}

\subsection{Error Analysis}
Examining the output from the fine-tuned GPT-3 model, we observe a number of recurring errors. 

The most notable errors are those where the model fails to link a mention to (all of) its coreferences that are present in the dialogue segment, or links mentions that denote different referents.
For example, for one mention the ground truth label is \textit{``the sheep dog''}, but the generated label was \textit{``the sheep dog with a leash''}; the model incorrectly attributed the prepositional phrase to the mention as it was actually a descriptor for a different referent. 
Related, since the CRDGs function at the message level, a mention can have both anaphoric and cataphoric coreferences when there are multiple mentions of the same referent in an utterance
An example of such an utterance is \textit{``Good question. I think the angry one also looks a little wild. So that could be an option as well. I mean the one with white nose and forehead''}, where \textit{``the angry one''}, \textit{``that''}, and \textit{``the one with white nose and forehead''} are all mentions of the same referent with the same ground truth label \textit{``the angry dog with a white nose and forehead''}. The model generates this correctly for the latter two, but not the former one for which only \textit{``the angry dog''} was generated, meaning it correctly substituted the proform but did not link the mention with its cataphoric coreferences.

Finally, some generated referent descriptions differ from the ground truth in terms of lexical choice or syntax, but not in terms of information content. This negatively affects scores of text generation metrics based on overlapping content in particular, but these are otherwise not meaningful errors as there are multiple ways to construct semantically similar descriptions, e.g., \textit{``the big dog which looks scary''} versus \textit{``the big scary-looking dog''}.

\section{Discussion}
We have presented an approach to visually-grounded reference resolution that frames the discourse processing side of the task as a causal language modeling problem. 
By fine-tuning an LLM to generate referent descriptions for marked mentions in dialogue segments from the collaborative image ranking task A Game Of Sorts \citep{willemsen_collecting_2022}, we demonstrate the possibility of treating referent identification as a zero-shot text-image retrieval problem by using pretrained VLMs for the grounding of the generated labels.
As we have not in any way indicated coreferential relations in the fine-tuning training data, our results imply that certain pretrained LLMs, here GPT-3, may learn to resolve coreferences implicitly without the need for explicit supervision for this fundamental subtask. 

In this work, we have treated the processing of the discourse as entirely disjoint of the visual modality. 
As such, it has inherent limitations. 
The mentions we find in the dialogues have not been produced void of the extralinguistic context. 
The dialogue participants could rely on co-observed visual stimuli to help resolve otherwise ambiguous language use.
From linguistic context alone, some ambiguities, such as prepositional phrase attachment, may be impossible to resolve. 
It is, therefore, noteworthy that the downstream zero-shot text-image retrieval performance using the generated descriptions from our unimodal approach far exceeds chance level accuracy, with the potential for results to improve further given access to the full dialogue history, as we found that the ground truth labels based on larger context windows achieve greater text-image retrieval performance. However, the results from our holistic human evaluation support the notion that a multimodal approach should ultimately prove even more effective.

We found that a decrease in text generation metric scores did not necessarily indicate a similar decrease in text-image retrieval performance, suggesting that the generated descriptions captured sufficiently discriminative information about the referents and achieved similar grounding accuracy despite not approximating the ground truth labels to the same extent.
It is also important to note that mentions may not have a single, canonical ground truth referent description due to lexical and syntactic variations between referring attempts.

Despite the relatively small size of the dataset collected by \citet{willemsen_collecting_2022}, we were still able to fine-tune GPT-3 to perform the task with greater accuracy than the baselines, which speaks to the sample efficiency of (certain) pretrained LLMs. In comparison, we find that the much smaller GPT-2 is prone to intrusions from the fine-tuning training data and more often fails to resolve the coreferences correctly.
Although the complexity of the discourse warrants the use of more powerful models, it is, nevertheless, likely that any LLM used for the task would benefit from a larger fine-tuning dataset. 
Related, benchmarking performance on other visually-grounded dialogue tasks would provide insights into the generalizability of the method. 

In addition to pursuing a multimodal approach, finer-grained grounding, and evaluating our method on other datasets, possible avenues for future work include expanding the annotations to include coreferential relations other than identity relations, addressing multi-image referents, and unifying the method with a mention proposal system.

\section*{Acknowledgements}
This work was partially supported by the Wallenberg AI, Autonomous Systems and Software Program (WASP) funded by the Knut and Alice Wallenberg Foundation. The authors would like to thank Erik Ekstedt, Dmytro Kalpakchi, Rajmund Nagy, Jim O'Regan, Ambika Kirkland, Chris Emmery, Chris van der Lee, and the anonymous reviewers for their helpful comments.

\bibliography{references}
\bibliographystyle{acl_natbib}

\appendix

\clearpage

\section*{Appendices}

\section{Dialogue Excerpt}
\label{sec:appendix-dialogue-excerpt}
\begin{figure}[h]
\begin{minipage}{1\textwidth}
    \centering
    \includegraphics[width=1\textwidth]{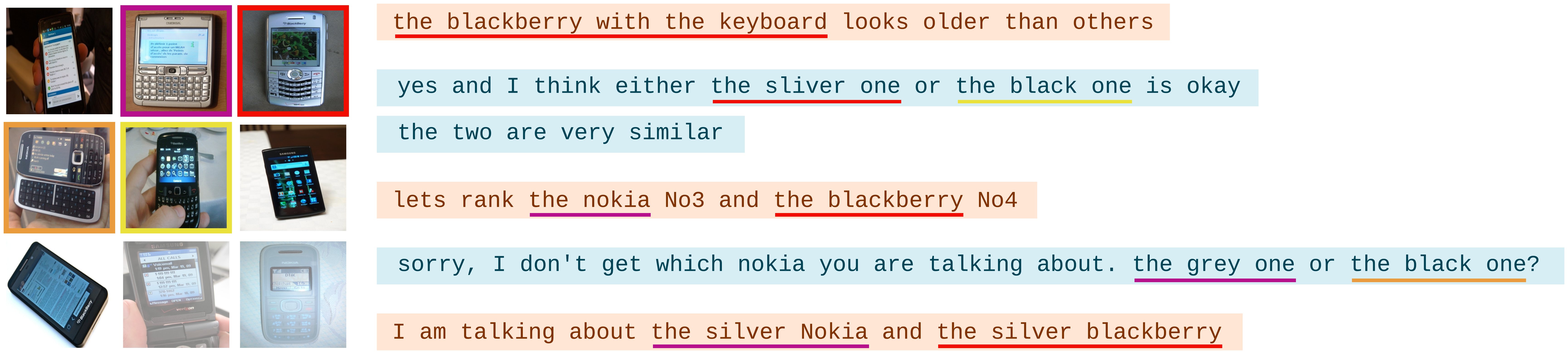}
    \caption{Excerpt of an AGOS dialogue with references to single-image referents underlined; the color indicates the referent. \textit{Note}. The two images that have been ranked successfully at this point in the interaction have a faded appearance.}
    \label{fig:dialogue-excerpt}
\end{minipage}
\end{figure}

\section{Model Input}
\label{sec:appendix-model-input}

\begin{figure}[h]
\begin{minipage}{\textwidth}
    \centering
    \includegraphics[width=1\textwidth]{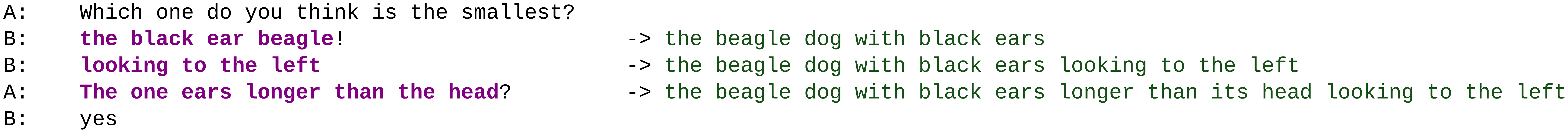}
    \caption{Excerpt of an AGOS dialogue with messages paired with manually constructed ground truth referent descriptions. Mentions are in purple and made bold for illustrative purposes. Ground truth labels for the referent denoted by the mention in green. \\\\}
    \label{fig:label_example}
\end{minipage}
\begin{minipage}{\textwidth}
    \centering
    \includegraphics[width=1\textwidth]{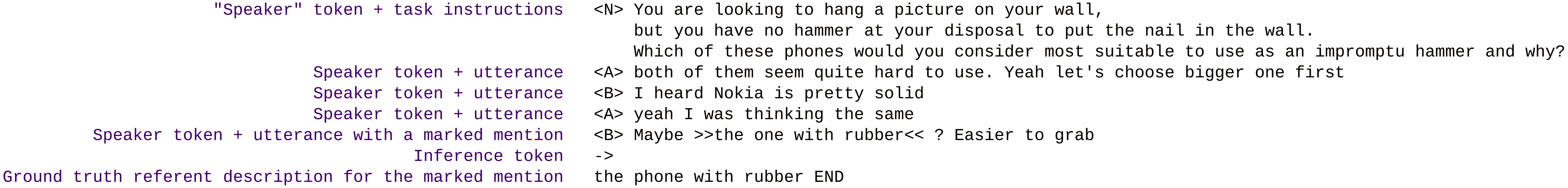}
    \caption{Sample input to LLM, deconstructed for demonstration purposes (the sample is otherwise a flat sequence of tokens). Left (text in purple): explanation of input; right (text in black): input. \textit{Note}. The ground truth is only available to the model during training, not during inference.}
    \label{fig:model_input}
\end{minipage}

\end{figure}

\section{Additional VLM Results}
\label{sec:appendix-vlm-results}

\begin{table}[h!]
\begin{minipage}{\textwidth}
    \footnotesize
    \centering
    \begin{tabular}{lcc|cc|cc|cc}
    \hline
    \multicolumn{1}{c}{} & \multicolumn{2}{c}{$\mathbf{CLIP}$\textbf{-}$\mathbf{B}$} & \multicolumn{2}{|c|}{$\mathbf{CLIP}$\textbf{-}$\mathbf{L}$} & \multicolumn{2}{|c|}{$\mathbf{ALIGN}$} & \multicolumn{2}{c}{$\mathbf{BLIP}$}  \\ 
    \cline{2-9}
    & $\mathbf{3}$ & $\mathbf{7}$  & $\mathbf{3}$ & $\mathbf{7}$  & $\mathbf{3}$ & $\mathbf{7}$  & $\mathbf{3}$ & $\mathbf{7}$  \\ 
    \hline
    Random & .11 & .11 & .11 & .11 & .11 & .11 & .11 & .11  \\ 
    Mention & .36 & .36 & .44 & .44 & .44 & .44 & .40 & .40 \\ 
    Substitution & .42 & .42 & .51 & .51 & .52 & .52 & .50 & .50 \\ 
    \hline
    coref, chain & .42 & .42 & .49 & .49 & .47 & .46 & .47 & .46 \\ 
    coref, set & .42 & .41 & .48 & .48 & .49 & .48 & .47 & .47 \\ 
    \hline
    GT, chain & .45 & .47 & .54 & .56 & .53 & .53 & .52 & .54 \\ 
    GT, set & .46 & .48 & .54 & .56 & .54 & .54 & .53 & .55 \\ 
    GT, manual & .47 & .48 & .53 & .55 & .58 & .59 & .55 & .57 \\ 
    \hline
    GPT-2 & .41 & .38 & .46 & .43 & .49 & .44 & .47 & .43 \\ 
    GPT-3 & .44 & .45 & .52 & .52 & .54 & .55 & .52 & .52 \\
    \hline
    \end{tabular}
\caption{\label{additional-ir-reduced-results}
Cross-validated image retrieval accuracy averaged over five folds for single-image referents (candidate set not reduced). \textit{Note}.
Scores are rounded to the nearest hundredth. GT = ground truth; CLIP-B = CLIP ViT-B/32; CLIP-L = CLIP ViT-L/14. \\
}
\end{minipage}
\end{table}

\clearpage

\begin{table}
\begin{minipage}{\textwidth}
    \footnotesize
    \centering
    \begin{tabular}{lcc|cc|cc|cc}
    \hline
    \multicolumn{1}{c}{} & \multicolumn{2}{c}{$\mathbf{CLIP}$\textbf{-}$\mathbf{B}$} & \multicolumn{2}{|c|}{$\mathbf{CLIP}$\textbf{-}$\mathbf{L}$} & \multicolumn{2}{|c|}{$\mathbf{ALIGN}$} & \multicolumn{2}{c}{$\mathbf{BLIP}$}  \\ 
    \cline{2-9}
    & $\mathbf{3}$ & $\mathbf{7}$  & $\mathbf{3}$ & $\mathbf{7}$  & $\mathbf{3}$ & $\mathbf{7}$  & $\mathbf{3}$ & $\mathbf{7}$  \\ 
    \hline
    Random & .22 & .22 & .22 & .22 & .22 & .22 & .22 & .22  \\ 
    Mention & .49 & .49 & .55 & .55 & .56 & .56 & .54 & .54 \\ 
    Substitution & .56 & .56 & .62 & .62 & .64 & .64 & .64 & .64 \\ 
    \hline
    coref, chain & .54 & .54 & .61 & .61 & .60 & .60 & .61 & .61 \\ 
    coref, set & .54 & .53 & .60 & .60 & .61 & .61 & .61 & .61 \\ 
    \hline
    GT, chain & .58 & .59 & .66 & .67 & .66 & .67 & .66 & .68 \\ 
    GT, set & .58 & .60 & .66 & .68 & .67 & .67 & .66 & .69 \\ 
    GT, manual & .59 & .60 & .64 & .66 & .69 & .70 & .69 & .70 \\ 
    \hline
    GPT-2 & .53 & .49 & .58 & .54 & .61 & .58 & .60 & .58 \\ 
    GPT-3 & .57 & .58 & .63 & .63 & .66 & .66 & .67 & .67 \\
    \hline
    \end{tabular}
\caption{\label{additional-ir-results}
Cross-validated image retrieval accuracy averaged over five folds for single-image referents (candidate set reduced). \textit{Note}. 
Scores are rounded to the nearest hundredth. GT = ground truth; CLIP-B = CLIP ViT-B/32; CLIP-L = CLIP ViT-L/14. \\\\
}
\end{minipage}
\begin{minipage}{\textwidth}
    \footnotesize
    \centering
    \begin{tabular}{lcc|cc|cc|cc|cc}
    \hline
    \multicolumn{1}{c}{} & \multicolumn{2}{c}{Cars} & \multicolumn{2}{|c|}{Dogs} & \multicolumn{2}{|c|}{Paintings} & \multicolumn{2}{|c|}{Pastries} & \multicolumn{2}{|c}{Phones}   \\ 
    \cline{2-11}
    & $\mathbf{3}$ & $\mathbf{7}$  & $\mathbf{3}$ & $\mathbf{7}$  & $\mathbf{3}$ & $\mathbf{7}$  & $\mathbf{3}$ & $\mathbf{7}$  & $\mathbf{3}$ & $\mathbf{7}$  \\ 
    \hline
    Random & .22 & .22 & .22 & .22 & .22 & .22 & .22 & .22 & .22 & .22  \\ 
    Mention & .52 & .52 & .62 & .62 & .60 & .60 & .61 & .61 & .58 & .58   \\ 
    Substitution & .63 & .63 & .70 & .70 & .70 & .70 & .68 & .68 & .67 & .67   \\ 
    \hline
    coref, chain & .59 & .60 & .69 & .69 & .66 & .67 & .67 & .68 & .63 & .63   \\ 
    coref, set & .60 & .57 & .68 & .68 & .69 & .68 & .69 & .70 & .62 & .62   \\ 
    \hline
    GT, chain & .66 & .66 & .76 & .78 & .72 & .74 & .75 & .78 & .71 & .69   \\ 
    GT, set & .66 & .65 & .74 & .77 & .73 & .78 & .76 & .80 & .73 & .73   \\ 
    GT, manual & .64 & .63 & .75 & .78 & .77 & .80 & .70 & .72 & .74 & .74   \\ 
    \hline
    GPT-2 & .62 & .62 & .67 & .62 & .67 & .62 & .63 & .61 & .57 & .50   \\ 
    GPT-3 & .63 & .63 & .75 & .78 & .70 & .70 & .68 & .72 & .70 & .69   \\
    \hline
    \end{tabular}
\caption{\label{additional-ir-results-per-fold}
Cross-validated image retrieval accuracy per fold for single-image referents (candidate set reduced). \textit{Note}. Scores shown are of VLM that averaged best performance (BLIP-2). Scores are rounded to the nearest hundredth. GT = ground truth. \\\\
}
\end{minipage}
\end{table}

\begin{table}[h!]
\footnotesize
\centering
\begin{tabular}{lcc|cc|cc}
\hline
\multicolumn{1}{c}{} & \multicolumn{2}{c}{$\mathbf{Accuracy}$} & \multicolumn{2}{|c|}{$\mathbf{MRR}$} & \multicolumn{2}{c}{$\mathbf{NDCG}$} \\ 
\cline{2-7}
& $\mathbf{3}$ & $\mathbf{7}$  & $\mathbf{3}$ & $\mathbf{7}$  & $\mathbf{3}$ & $\mathbf{7}$  \\ 
\hline
Random & .11 & .11  & .22 & .22 & - & - \\ 
Mention & .47 & .47 & .63 & .63 & .72 & .72 \\ 
Substitution & .55 & .55 & .71 & .71 & .78 & .78 \\ 
\hline
coref, chain & .53 & .51 & .69 & .68 & .76 & .76 \\ 
coref, set & .53 & .51 & .69 & .68 & .77 & .76 \\ 
\hline
GT, chain & .60 & .61 & .75 & .76 & .81 & .82 \\ 
GT, set & .60 & .62 & .75 & .77 & .81 & .83 \\ 
GT, manual & .63 & .64 & .76 & .78 & .82 & .83 \\ 
\hline
GPT-2 & .54 & .48 & .69 & .65 & .77 & .73 \\ 
GPT-3 & .60 & .60 & .74 & .74 & .80 & .81 \\
\hline
\end{tabular}
\caption{\label{main-ir-reduced-results}
Cross-validated image retrieval performance averaged over five folds for single-image referents (candidate set not reduced). \textit{Note}. 
Scores shown are of VLM that averaged best performance (BLIP-2).
Scores are rounded to the nearest hundredth. GT = ground truth.
}
\end{table}

\newpage \strut 

\begin{figure}[b!]
    \centering
    \includegraphics[width=1\columnwidth]{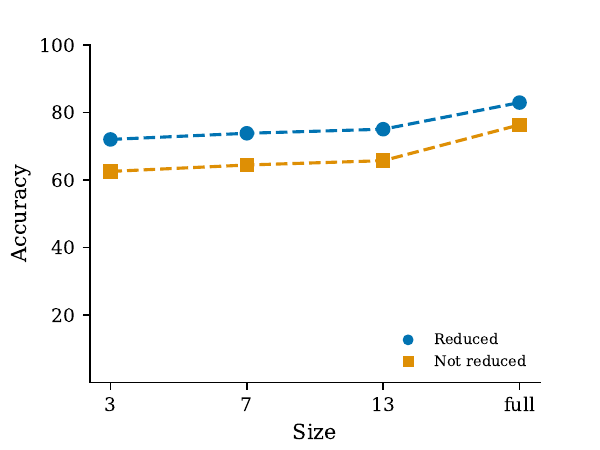}
    \caption{Text-image retrieval accuracy as a function of the size of the context window. Results are shown for BLIP-2 on the manually constructed ground truth referent descriptions based on their respective windows. We show results for both the reduced candidate set and the not reduced candidate set.}
    \label{fig:ground_truth_accuracy}
\vspace{3.25cm}
\end{figure}

\end{document}